\documentclass{article}
\usepackage[T1]{fontenc}
\usepackage{authblk}
\usepackage{graphicx}
\usepackage{hyperref}
\usepackage{amsmath}
\usepackage{booktabs}
\usepackage{multirow}
\usepackage{multicol}
\usepackage{cleveref}
\usepackage[dvipsnames]{xcolor}

\crefname{figure}{Fig.}{Figs.}
\crefname{table}{Table}{Tables}
\crefname{section}{Section}{Sections}
\crefname{subsection}{Section}{Sections}
\crefname{equation}{}{}

\newcommand{\ourmethod}{CLARITY}
\newcommand{\nlabeled}{97}

\begin{document}

\title{A Clinical Evaluation of Conformal Prediction for Chest Radiograph Report Generation}

\author[1]{Nikolaos Kontogeorgis\thanks{Corresponding author: \href{mailto:nikolaos.kontogeorgis@proton.me}{nikolaos.kontogeorgis@proton.me}.}}
\author[2]{Pascal Krezer}
\author[1]{Klaus-Rudolf Kladny}
\author[2]{Arne Estler}
\author[3]{Christian F. Baumgartner}
\author[4]{Lisa Koch}
\author[2]{Saif Afat}
\author[1,5]{Michael Muehlebach}
\affil[1]{Max Planck Institute for Intelligent Systems, T\"ubingen, Germany}
\affil[2]{University Hospital T\"ubingen, T\"ubingen, Germany}
\affil[3]{University of Lucerne, Lucerne, Switzerland}
\affil[4]{University of Bern, Bern, Switzerland}
\affil[5]{Vienna University of Economics and Business}
\renewcommand\Authand{ and }
\renewcommand\Authands{ and }

\date{}

\maketitle

\begin{abstract}
	While multimodal large language models (MLLMs) have improved automated radiology report generation, they remain prone to hallucinations, necessitating clinically meaningful risk-control methods. Conformal prediction (CP) provides statistical guarantees, yet existing CP approaches rely on uncalibrated automated extractors and fail to account for the clinical relevance of errors. We introduce \underline{cl}inical \underline{a}dmissibility c\underline{rit}eria (\ourmethod{}), a framework for evaluating radiology reports that explicitly distinguishes clinically consequential errors from minor deviations. To eliminate reliance on automatic judges, we collected radiologist annotations of MLLM-generated reports using \ourmethod{}. By benchmarking CP strategies and automatic evaluation metrics against these expert annotations, we demonstrate that commonly used admissibility proxies systematically fail to detect clinically relevant errors. Our findings reveal that automated admissibility judgments are largely decoupled from true clinical admissibility.
\end{abstract}

\section{Introduction}

\begin{figure}[t]
	\centering
	\includegraphics[width=0.99\textwidth]{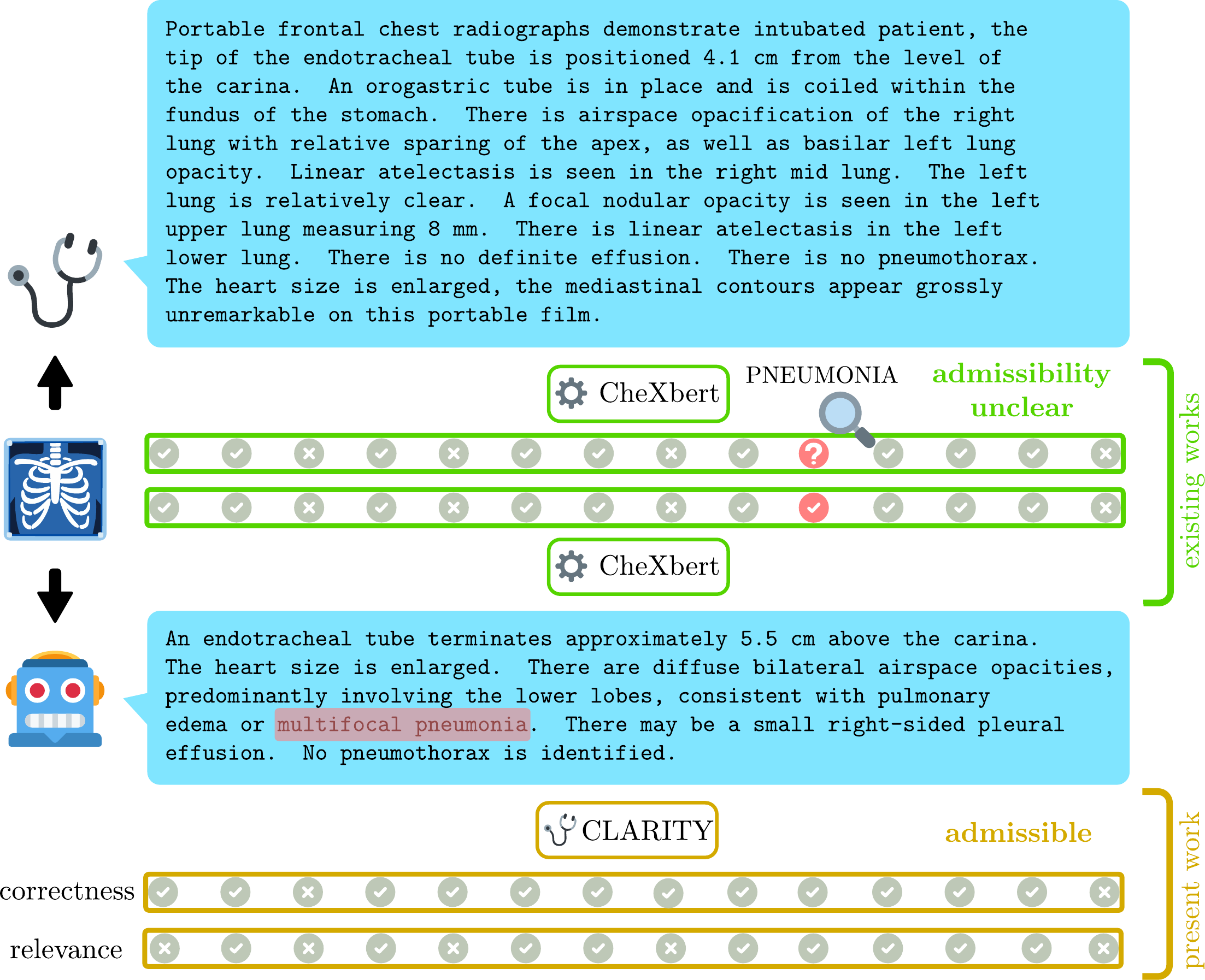}
	\caption{\textbf{Qualitative comparison of admissibility assessments.} Reference-based automatic assessment maps a candidate and reference report to clinical labels and then compares those labels. This procedure can penalize a correct finding that was omitted from the reference and treats discrepancies equally without accounting for their clinical consequence. \ourmethod{} instead uses image-grounded radiologist judgments of both relevance and accuracy. The labels and arrows, rather than color alone, distinguish the assessment pathways.}
	\label{fig:main}
\end{figure}

Machine learning has achieved striking performance in medical applications ranging from breast cancer screening~\cite{laang2023artificial} to cardiovascular risk prediction~\cite{lubeck2025adaptable}. Automated radiology report generation has progressed rapidly~\cite{li2025automatic}, yet safe clinical deployment requires trustworthy uncertainty estimates. Miscalibration and hallucination remain key failure modes~\cite{van2019calibration,kim2025medical}: overconfident or fabricated claims can directly mislead clinical decisions~\cite{asgari2025framework}. Conformal prediction (CP) addresses this problem by producing sets of candidate reports with finite-sample guarantees that at least one admissible report is included with a certain target probability~\cite{clm,scope-gen}.

For both calibration and evaluation, CP methods require an \textit{admissibility function} that determines whether a generated report is acceptable. Existing approaches use proxies such as linguistic metrics, including BLEU~\cite{papineni2002bleu} and ROUGE~\cite{lin2004rouge}, or automated clinical extractors such as CheXbert~\cite{chexbert} and RadGraph~\cite{jain2021radgraph}. Linguistic metrics are sensitive to wording and do not directly encode clinical correctness, whereas learned extractors inherit model and reference-report errors and do not represent the asymmetric consequences of different mistakes. Encoding generated and reference reports as label sets also fails to capture the clinical relevance of individual findings or errors in the reference report itself.

Evaluation research has consequently moved toward embedding similarity, structured clinical extraction, and domain-specific factuality measures. Among these approaches, BERTScore compares contextual representations~\cite{bertscore}; RadCliQ combines report metrics to better reflect radiologist error assessments~\cite{yu2023evaluating}; GREEN identifies clinically significant errors in candidate reports~\cite{ostmeier2024green}; and RaTE-Score and FORTE emphasize clinical entities and structured radiology concepts~\cite{zhao2024ratescore,li2025forte}. Safety-oriented generation methods have similarly included grounding statements to image evidence~\cite{maira-2}, black-box uncertainty quantification~\cite{scuq}, and conformal methods for language and vision-language models~\cite{clm,scope-gen,shahrokhi2025conformal,conflvlm}. Any conformal guarantee, however, remains defined relative to its chosen admissibility event.

We address this gap by introducing \ourmethod{}, a radiologist-grounded framework that distinguishes clinically consequential errors from minor omissions and stylistic differences. We use \ourmethod{} to benchmark CP methods and quantify the extent to which automatic metrics align with the resulting operational definition of clinical admissibility.

Our main contributions are:
\begin{enumerate}
	\item \textbf{Clinically grounded admissibility criteria.} We introduce \ourmethod{}, a radiologist-grounded framework that assesses report admissibility by distinguishing clinically relevant errors from minor omissions and stylistic differences.
	\item \textbf{Expert-annotated dataset.} We obtain board-certified radiologist annotations for reports generated by a MLLM, releasing the dataset for reproducible benchmarking.
	\item \textbf{Misalignment of automatic metrics with radiologist evaluation.} We systematically compare 14 automatic metric variants with \ourmethod{} and show that their alignment with radiologist-assessed clinical admissibility is metric-dependent and incomplete. In particular, CheXbert positive-label F1, a commonly used admissibility proxy, shows no positive alignment with the radiologist endpoint, while even the strongest metric is not an interchangeable substitute for expert evaluation.
	\item \textbf{CP benchmarking.} We systematically compare CP methods under clinically grounded evaluation, including unsuccessful calibration runs, and characterize the coverage--set-size trade-offs observed under different admissibility definitions.
\end{enumerate}

\section{Results} \label{sec:experiments}
Our primary experiment tests agreement between existing automatic metrics and the radiologist assessments defined by \ourmethod{}. We then examine the downstream consequences of disagreement by calibrating and evaluating conformal prediction with either CheXbert or \ourmethod{} admissibility.

\subsection{Misalignment of Existing Metrics with \ourmethod{}} \label{sec:qualitative_comparison}

\begin{figure}[t]
	\centering
	\includegraphics[width=0.85\textwidth]{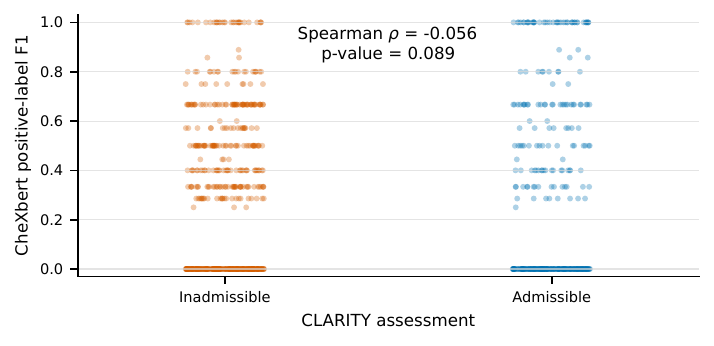}
	\caption{\textbf{Comparison of CheXbert and \ourmethod{}.} Distribution of CheXbert positive-label F1 scores for 970 generated reports from 97 studies, grouped by radiologist-assessed \ourmethod{} admissibility. Points show individual reports.}
	\label{fig:chexbert_vs_clarity}
\end{figure}

\begin{table}[t]
	\centering
	\fontsize{8pt}{9.6pt}\selectfont
	\caption{Association of automatic metrics with \ourmethod{}. Spearman correlations are computed after study-level averaging; 95\% confidence intervals use 2,000 study-cluster bootstrap samples and two-sided \(P\) values use Holm correction across 14 metric variants. RadCliQ scores are sign-inverted so higher values consistently indicate better reports. The sample-size column gives studies/reports after metric-specific exclusion of undefined values.}
	\label{tab:metric_corr}
	\setlength{\tabcolsep}{5pt}
	\resizebox{\linewidth}{!}{%
		\begin{tabular}{lcccc}
			\toprule
			Metric                                      & \(n\) & \(\rho\) & 95\% CI         & \(P_{\mathrm{Holm}}\)      \\
			\midrule
			GREEN~\cite{ostmeier2024green}              & 97/970 & 0.700    & [0.562, 0.798]  & \(2.05\!\times\!10^{-14}\) \\
			\(-\)RadCliQ-v1~\cite{yu2023evaluating}     & 97/970 & 0.479    & [0.295, 0.630]  & \(8.87\!\times\!10^{-6}\)  \\
			\(-\)RadCliQ-v0~\cite{yu2023evaluating}     & 97/970 & 0.476    & [0.288, 0.630]  & \(9.77\!\times\!10^{-6}\)  \\
			BERTScore~\cite{bertscore}                  & 97/970 & 0.467    & [0.282, 0.624]  & \(1.61\!\times\!10^{-5}\)  \\
			FORTE pooled F1 (custom)~\cite{li2025forte} & 97/970 & 0.432    & [0.247, 0.586]  & \(9.69\!\times\!10^{-5}\)  \\
			RaTE-Score~\cite{zhao2024ratescore}         & 97/970 & 0.402    & [0.212, 0.572]  & \(4.02\!\times\!10^{-4}\)  \\
			semb-score~\cite{yu2023evaluating}          & 97/970 & 0.386    & [0.181, 0.565]  & \(7.71\!\times\!10^{-4}\)  \\
			FORTE degree F1~\cite{li2025forte}          & 97/970 & 0.383    & [0.208, 0.536]  & \(7.71\!\times\!10^{-4}\)  \\
			RadGraph-F1~\cite{jain2021radgraph}         & 97/970 & 0.372    & [0.180, 0.540]  & 0.00103                    \\
			BLEU~\cite{papineni2002bleu}                & 97/970 & 0.273    & [0.060, 0.461]  & 0.0339                     \\
			FORTE impression F1~\cite{li2025forte}      & 96/957 & 0.235    & [0.016, 0.445]  & 0.0854                     \\
			FORTE feature F1~\cite{li2025forte}         & 97/956 & 0.229    & [0.026, 0.399]  & 0.0854                     \\
			FORTE landmark F1~\cite{li2025forte}        & 97/925 & 0.135    & [-0.068, 0.310] & 0.373                      \\
			CheXbert positive F1~\cite{chexbert}        & 97/970 & -0.126   & [-0.339, 0.093] & 0.373                      \\
			\bottomrule
	\end{tabular}}
\end{table}

The automatic metrics did not uniformly recover radiologist-assessed admissibility. An empirical exploratory analysis shows that report-level CheXbert positive-label F1 was weakly negatively associated with \ourmethod{} (\(\rho=-0.070\), unadjusted two-sided \(P=0.030\)). The primary study-level association was also negative and was not significant after multiplicity correction (\(\rho=-0.126\), 95\% CI -0.339--0.093; Holm-adjusted \(P=0.373\)). CheXbert therefore provided no evidence of positive alignment with the radiologist endpoint despite its use as an admissibility proxy in prior CP pipelines (\cref{fig:chexbert_vs_clarity}).

As shown in \cref{tab:metric_corr}, the degree of alignment varied substantially across metrics. GREEN had the strongest study-level association (\(\rho=0.700\), 95\% CI 0.562--0.798; Holm-adjusted \(P=2.05\!\times\!10^{-14}\)), plausibly because it penalizes clinically significant errors while discounting insignificant ones, thereby most closely reflecting the \ourmethod{} relevance dimension. RadCliQ, BERTScore, custom pooled FORTE, RaTE-Score, semb-score, FORTE degree F1, RadGraph F1, and BLEU also had positive associations with Holm-adjusted \(P<0.05\). By contrast, the FORTE impression, feature, and landmark components were not significant after correction, and CheXbert positive F1 was negatively correlated. Among the FORTE variants, the custom pooled score had the strongest association (\(\rho=0.432\)), followed by \textsc{degree} (\(\rho=0.383\)), \textsc{impression} (\(\rho=0.235\)), \textsc{feature} (\(\rho=0.229\)), and \textsc{landmark} (\(\rho=0.135\)). At report level, 39.4\% of generated reports were \ourmethod{}-admissible. Automatic metrics therefore captured useful aggregate signal to markedly different degrees, but even the strongest association remained imperfect; none was an interchangeable substitute for the radiologist-grounded endpoint.

\subsection{Application to Conformal Prediction}\label{sec:eval}
\begin{table}[t]
	\centering
	\fontsize{8pt}{9.6pt}\selectfont
	\caption{Empirical coverage and prediction-set size for conformal miscoverage \(\alpha=0.30\). Random-split columns report the mean~\(\pm\)~standard deviation (s.d.) over the finite runs among 100 attempted study-level splits and the number of finite runs. The fixed disjoint CheXbert-to-\ourmethod{} evaluation uses the primary F1 threshold of 0.4 and reports one run.}
	\label{tab:results0.3}
	\setlength{\tabcolsep}{5pt}
	\resizebox{\linewidth}{!}{%
		\begin{tabular}{llcccc}
			\toprule
			\textbf{Method} & \textbf{Metric}            & \textbf{CheXbert (full set)} & \textbf{CheXbert (disjoint \ourmethod{} eval.)} & \textbf{CheXbert (labeled set)} & \textbf{\ourmethod{}} \\
			\midrule
			& \(n_{\text{calibration}}\) & 234                          & 196                                             & 77                              & 77                    \\
			& \(n_{\text{test}}\)        & 59                           & 97                                              & 20                              & 20                    \\
			\midrule
			\multirow{3}{*}{SCOPE-Gen (count)}
			& Coverage                   & \(0.72 \pm 0.06\)            & \(0.67\)                                        & \(0.72 \pm 0.11\)               & \(0.73 \pm 0.11\)     \\
			& Pred.\ set size            & \(5.63 \pm 0.58\)            & \(5.00\)                                        & \(6.56 \pm 1.27\)               & \(8.63 \pm 1.67\)     \\
			& \(n_{\text{valid}}\)       & 100                          & 1                                               & 100                             & 100                   \\
			\midrule
			\multirow{3}{*}{SCOPE-Gen (max)}
			& Coverage                   & \(0.71 \pm 0.06\)            & \(0.68\)                                        & \(0.69 \pm 0.09\)               & \(0.62 \pm 0.07\)     \\
			& Pred.\ set size            & \(4.94 \pm 0.68\)            & \(4.70\)                                        & \(5.88 \pm 0.97\)               & \(5.50 \pm 1.10\)     \\
			& \(n_{\text{valid}}\)       & 100                          & 1                                               & 90                              & 31                    \\
			\midrule
			\multirow{3}{*}{CLM (count)}
			& Coverage                   & \(0.77 \pm 0.04\)            & \(0.75\)                                        & \(0.62 \pm 0.03\)               & \(0.64 \pm 0.06\)     \\
			& Pred.\ set size            & \(9.43 \pm 0.56\)            & \(9.59\)                                        & \(9.12 \pm 0.88\)               & \(8.91 \pm 0.53\)     \\
			& \(n_{\text{valid}}\)       & 44                           & 1                                               & 2                               & 6                     \\
			\midrule
			\multirow{3}{*}{CLM (max)}
			& Coverage                   & \(0.75 \pm 0.06\)            & \(0.76\)                                        & \(0.71 \pm 0.05\)               & \(0.65 \pm 0.08\)     \\
			& Pred.\ set size            & \(8.98 \pm 0.78\)            & \(9.91\)                                        & \(8.43 \pm 0.47\)               & \(8.02 \pm 0.42\)     \\
			& \(n_{\text{valid}}\)       & 43                           & 1                                               & 4                               & 3                     \\
			\bottomrule
	\end{tabular}}
\end{table}

At the primary CheXbert-F1 threshold of 0.4, all fixed disjoint configurations produce finite sets. At \(\alpha=0.3\), CLM exceeds the 0.70 coverage target but requires 9.59--9.91 of ten reports, whereas SCOPE-Gen produces smaller sets of 4.70--5.00 reports but reaches only 0.67--0.68 coverage. At \(\alpha=0.45\), CLM again exceeds the 0.55 target with sets of 8.05--10.00 reports; SCOPE-Gen returns sets of 2.00--2.23 reports but reaches only 0.53--0.54 coverage. Thus, transfer from CheXbert calibration to \ourmethod{} evaluation exposes a selectivity--coverage trade-off rather than uniformly preserving the nominal target.

The sensitivity analysis in \cref{tab:disjoint_sensitivity} shows that this conclusion depends materially on the CheXbert-F1 threshold. Raising the threshold from 0.4 to 0.5 generally enlarges the SCOPE-Gen sets; at \(\alpha=0.3\), both CLM configurations become invalid. At 0.6, no method produces a finite set for \(\alpha=0.3\), while the finite \(\alpha=0.45\) sets contain 5.42--9.91 reports. Threshold selection therefore changes both feasibility and practical utility and should be treated as part of the admissibility definition rather than as a benign tuning choice. A similar low-selectivity pattern appears under \ourmethod{} calibration: at \(\alpha=0.45\), CLM returns 7.80--7.84 reports on average, compared with 3.05--3.07 for SCOPE-Gen.

Across the random-split experiments, SCOPE-Gen generally yields smaller prediction sets than CLM. This advantage must be interpreted together with calibration stability. At \(\alpha=0.3\), some CLM summaries are based on only two to six valid runs, and the SCOPE-Gen \(\mathrm{max}\) configuration under \ourmethod{} is based on 31 valid runs. Means computed from these successful subsets can be optimistic and do not describe all 100 attempted splits. At \(\alpha=0.45\), SCOPE-Gen is stable across all random splits, whereas CLM remains invalid in 45--52\% of labeled-set runs. Collectively, the results show that coverage, prediction-set size, and calibration failure must be considered jointly; an admissibility proxy that yields nominal coverage can still be impractical or fail to transfer to clinically grounded evaluation.

\section{Discussion} \label{sec:discussion}
The central finding is that existing automatic metrics recover radiologist-assessed clinical admissibility to markedly different and incomplete degrees. GREEN shows by far the strongest association with \ourmethod{} (\(\rho=0.700\)), plausibly because its error counts target clinically significant errors while discounting insignificant ones---the design that most closely mirrors the \ourmethod{} relevance dimension. RadCliQ, BERTScore, custom pooled FORTE, RaTE-Score, semb-score, FORTE degree F1, RadGraph-F1, and BLEU also capture meaningful study-level variation. However, the component-level FORTE results are heterogeneous, and CheXbert positive-label F1 shows no positive association with \ourmethod{}. This result is especially consequential because CheXbert has been used to define admissibility in prior conformal report-generation pipelines.

The mismatch reflects two structural limitations rather than merely a difference in metric sensitivity. First, as depicted in~\cref{fig:main}, reference-based automatic labelers are bound by the reference text: if a generated report correctly identifies a finding omitted by the original radiologist, CheXbert can penalize it as a hallucination. This limitation is consequential because expert radiologists found 30/97 (30.9\%) ground-truth reports in our dataset to be clinically inadmissible. Second, label-matching metrics ignore asymmetric clinical consequences. For example, failing to mention a pneumothorax and hallucinating a minor bone change each count as a single error under CheXbert, whereas \ourmethod{} uses its \textit{relevance} dimension to distinguish errors that could affect patient management. GREEN's comparatively strong association supports the importance of explicitly modeling clinical significance, but even GREEN leaves a gap to the radiologist endpoint. Automatic metrics may therefore remain useful for scalable model development, ranking, or cohort-level monitoring, while the present evidence does not support treating any single metric as an interchangeable replacement for image-grounded radiologist evaluation.

\begin{table}[t]
	\centering
	\fontsize{8pt}{9.6pt}\selectfont
	\caption{Empirical coverage and prediction-set size for conformal miscoverage \(\alpha=0.45\). Random-split columns report the mean~\(\pm\)~standard deviation (s.d.) over the finite runs among 100 attempted study-level splits and the number of finite runs. The fixed disjoint CheXbert-to-\ourmethod{} column uses the primary F1 threshold of 0.4 and reports one run.}
	\label{tab:results0.45}
	\setlength{\tabcolsep}{5pt}
	\resizebox{\linewidth}{!}{%
		\begin{tabular}{llcccc}
			\toprule
			\textbf{Method} & \textbf{Metric}            & \textbf{CheXbert (full set)} & \textbf{CheXbert (disjoint \ourmethod{} eval.)} & \textbf{CheXbert (labeled set)} & \textbf{\ourmethod{}} \\
			\midrule
			& \(n_{\text{calibration}}\) & 234                          & 196                                             & 77                              & 77                    \\
			& \(n_{\text{test}}\)        & 59                           & 97                                              & 20                              & 20                    \\
			\midrule
			\multirow{3}{*}{SCOPE-Gen (count)}
			& Coverage                   & \(0.65 \pm 0.06\)            & \(0.53\)                                        & \(0.61 \pm 0.10\)               & \(0.60 \pm 0.10\)     \\
			& Pred.\ set size            & \(2.98 \pm 0.14\)            & \(2.00\)                                        & \(3.04 \pm 0.20\)               & \(3.05 \pm 0.36\)     \\
			& \(n_{\text{valid}}\)       & 100                          & 1                                               & 100                             & 100                   \\
			\midrule
			\multirow{3}{*}{SCOPE-Gen (max)}
			& Coverage                   & \(0.57 \pm 0.08\)            & \(0.54\)                                        & \(0.58 \pm 0.11\)               & \(0.56 \pm 0.11\)     \\
			& Pred.\ set size            & \(2.40 \pm 0.21\)            & \(2.23\)                                        & \(3.20 \pm 0.35\)               & \(3.07 \pm 0.52\)     \\
			& \(n_{\text{valid}}\)       & 100                          & 1                                               & 100                             & 100                   \\
			\midrule
			\multirow{3}{*}{CLM (count)}
			& Coverage                   & \(0.78 \pm 0.05\)            & \(0.77\)                                        & \(0.72 \pm 0.10\)               & \(0.72 \pm 0.10\)     \\
			& Pred.\ set size            & \(9.91 \pm 0.36\)            & \(10.00\)                                       & \(8.70 \pm 1.58\)               & \(7.84 \pm 1.01\)     \\
			& \(n_{\text{valid}}\)       & 100                          & 1                                               & 48                              & 48                    \\
			\midrule
			\multirow{3}{*}{CLM (max)}
			& Coverage                   & \(0.74 \pm 0.07\)            & \(0.67\)                                        & \(0.67 \pm 0.08\)               & \(0.70 \pm 0.09\)     \\
			& Pred.\ set size            & \(8.94 \pm 0.69\)            & \(8.05\)                                        & \(6.80 \pm 1.12\)               & \(7.80 \pm 1.65\)     \\
			& \(n_{\text{valid}}\)       & 100                          & 1                                               & 50                              & 55                    \\
			\bottomrule
	\end{tabular}}
\end{table}

The conformal prediction experiments demonstrate that admissibility is not a neutral implementational detail: changing the criterion used to identify an acceptable report changes both the frequency of admissible calibration examples and the resulting coverage--set-size trade-off. Under the fixed disjoint configuration at the CheXbert-F1 threshold of 0.4, finite sets are possible at both target coverages, but SCOPE-Gen falls slightly below the nominal targets while CLM attains them using much larger sets. Increasing the threshold enlarges sets or makes calibration infeasible, culminating in no finite \(\alpha=0.3\) configuration at F1~\(\geq0.6\). Thus, the metric mismatch propagates into the feasibility and practical utility of proxy-calibrated prediction sets.

Within this application, SCOPE-Gen generally produces smaller sets than CLM, particularly at \(\alpha=0.45\), but method comparisons cannot be reduced to average set size. Several CLM configurations, and some SCOPE-Gen configurations at \(\alpha=0.3\), fail to produce finite sets for substantial fractions of the random splits. Moreover, the reported means are conditional on successful runs. This is particularly important where only two to six of 100 CLM runs are valid: those entries provide little evidence about expected behavior under a new split. Reporting \(n_{\text{valid}}\) alongside coverage and set size exposes this failure mode, while the large finite sets distinguish nominal coverage from practical utility.

\ourmethod{} addresses two limitations of reference-based admissibility. First, readers assess reports against the images rather than treating the original report as an error-free gold standard; this matters because 30 of 97 original reports are themselves rated inadmissible under the rubric. Second, the relevance label makes the admissibility event asymmetric: only findings judged relevant are required to be correct. This better reflects the fact that reporting errors differ in clinical importance, while remaining a transparent operational rule that can be audited at the category level. The rule should be viewed as an evaluation framework rather than a complete model of downstream harm. Future work could adapt category definitions or relevance thresholds to particular clinical workflows and should test whether those adaptations preserve calibration across institutions and reader populations.

\begin{table}[t]
	\centering
	\fontsize{8pt}{9.6pt}\selectfont
	\caption{Sensitivity of the fixed disjoint CheXbert-to-\ourmethod{} evaluation to the CheXbert positive-label F1 admissibility threshold. Each cell reports coverage / prediction-set size for the single fixed split; ``--'' indicates that no finite prediction set could be calibrated.}
	\label{tab:disjoint_sensitivity}
	\setlength{\tabcolsep}{5pt}
	\begin{tabular}{cllccc}
		\toprule
		\(\alpha\) & \textbf{Method} & \textbf{Score} & \(\mathrm{F1}\geq0.4\) & \(\mathrm{F1}\geq0.5\) & \(\mathrm{F1}\geq0.6\) \\
		\midrule
		\multirow{4}{*}{0.30}
		& SCOPE-Gen       & count          & 0.67 / 5.00            & 0.71 / 7.00            & --                     \\
		& SCOPE-Gen       & max            & 0.68 / 4.70            & 0.71 / 6.41            & --                     \\
		& CLM             & count          & 0.75 / 9.59            & --                     & --                     \\
		& CLM             & max            & 0.76 / 9.91            & --                     & --                     \\
		\midrule
		\multirow{4}{*}{0.45}
		& SCOPE-Gen       & count          & 0.53 / 2.00            & 0.60 / 3.00            & 0.70 / 6.00            \\
		& SCOPE-Gen       & max            & 0.54 / 2.23            & 0.54 / 2.44            & 0.71 / 5.42            \\
		& CLM             & count          & 0.77 / 10.00           & 0.77 / 10.00           & 0.74 / 9.16            \\
		& CLM             & max            & 0.67 / 8.05            & 0.69 / 6.88            & 0.76 / 9.91            \\
		\bottomrule
	\end{tabular}
\end{table}

The primary metric finding suggests a division of labor between automated and expert evaluation. Automatic metrics can support large-scale screening and identify broad changes in report quality, while smaller image-grounded expert datasets can determine whether each proxy remains aligned with the clinical endpoint of interest. Such validation is necessary before metric scores are interpreted as evidence of clinical admissibility, particularly when the metric reduces a report to lexical or extracted-label agreement.

The CP application demonstrates the downstream importance of this alignment check because conformal guarantees apply to the defined loss, not automatically to clinical safety. A deployment study should therefore report evidence that its admissibility function remains aligned with the target clinical endpoint, alongside nominal coverage, calibration failures, prediction-set size, and the prevalence of admissible candidates.

This study has several limitations. Only 97 of the 293 eligible studies were annotated, and the labeled cohort comprises the sequence of cases completed by the radiologists within the allotted annotation time rather than a randomized sample. Selection effects may therefore influence both admissibility prevalence and metric associations. The data come from a single public dataset, and all candidates were generated by one model variant with ten pre-generated reports per study. The findings may not generalize to other institutions, modalities, generators, prompts, or decoding procedures.

\ourmethod{} is also limited to its 12 categories and operational relevance judgments; clinically important phenomena outside the rubric may be missed. The metric analysis establishes association rather than interchangeability with the radiologist endpoint. Missing component-level FORTE values arise when no terms from a category are matched, producing unequal denominators and particularly sparse within-study information for the feature and landmark components, whereas GREEN covers all 970 reports in the final labeled cohort. The automatic metrics were not validated as replacement admissibility functions in otherwise identical CP pipelines.

Finally, the secondary CP application uses small calibration and test samples. Invalid conformal runs and the instability of conditional summaries at \(\alpha=0.3\) limit the precision of comparisons between methods, while the disjoint results represent one fixed split. Although the disjoint comparison uses the same primary CheXbert threshold as the random-split experiments, the sensitivity analysis confirms that feasibility and set size depend strongly on this threshold. Larger, prospectively sampled, multi-institutional studies with repeated independent annotations are needed before either the metric relationships or the conformal results can be interpreted as deployment-ready evidence.

We introduced \ourmethod{}, a radiologist-grounded operational definition of admissibility for automated chest-radiograph reports, and used it to test whether 14 automatic metric variants recover expert assessments of clinical admissibility. Our central finding is that alignment with radiologist evaluation is highly metric-dependent and incomplete. CheXbert positive-label F1, despite its use as an admissibility proxy in prior conformal pipelines, shows no positive alignment with the radiologist endpoint. GREEN aligns most strongly at study level, but even this association is imperfect and does not make the metric interchangeable with image-grounded expert review. The observed mismatch reflects limitations that label and text comparisons cannot resolve alone: reference reports may themselves contain errors, and superficially similar discrepancies can have substantially different clinical relevance.

We then examined how this mismatch propagates to conformal prediction. Changing the admissibility definition alters calibration feasibility, empirical coverage, and prediction-set size; SCOPE-Gen generally produces smaller sets than CLM, but can miss nominal coverage, whereas CLM often requires nearly all candidates or fails to calibrate on some splits. These results illustrate a practical consequence of the primary metric finding: statistical guarantees defined through an automatic proxy should not be interpreted as guarantees of radiologist-assessed clinical admissibility without direct validation. The expert-annotated benchmark provides a basis for more reproducible evaluation of metric alignment and clinically grounded risk control in radiology report generation.

\section{Methods}

\subsection{Study design and data source} \label{sec:dataset}
We conducted a retrospective methodological study using MIMIC-CXR~\cite{mimic-cxr}, a de-identified database containing 377,110 chest radiographs from 227,827 imaging studies with free-text reports. Because MedGemma was trained on the official MIMIC-CXR training and evaluation partitions, we restricted the analysis to the test partition. This partition contains 3,269 imaging studies from 293 patients. Multiple studies from the same patient would violate the patient-level exchangeability assumption underlying the conformal analyses~\cite{tibshirani2023conformal}; we therefore retained the chronologically first study for each patient, yielding 293 independent study units. The first study was selected by a prespecified deterministic rule rather than according to image or report content.

\subsection{Report generation} \label{sec:model}
We used the 27-billion-parameter instruction-tuned MedGemma medical vision-language model~\cite{medgemma}. The associated reports were divided into sections by rule-based parsing. Text from the \textsc{Indication} or \textsc{History} section was supplied as the clinical \textsc{Reason}, frontal and available lateral chest radiographs were supplied as image input, and the \textsc{Findings} section served as the reference report. A standardized prompt was used to generate ten candidate reports for each of the 293 studies. Preliminary inspection found the instruction-tuned checkpoint more coherent and precise than the pretrained checkpoint, and all subsequent analyses used the instruction-tuned variant.

\subsection{Clinical admissibility criteria} \label{sec:evaluation}
\ourmethod{} assesses a report for relevance as \textit{Relevant} or \textit{Irrelevant} and for accuracy as \textit{Correct}, \textit{Incorrect}, or \textit{Missing} in each of 12 clinical finding categories:

\begin{multicols}{2}
	\begin{enumerate}
		\item Diaphragm
		\item Pleural effusion
		\item Pneumothorax
		\item Mediastinum
		\item Heart silhouette
		\item Cardiopulmonary congestion
		\item Pneumonia
		\item Lung opacity
		\item Osseous structures
		\item Soft tissue
		\item Abdomen
		\item Foreign material
	\end{enumerate}
\end{multicols}

The categories are largely inspired by the popular CheXpert labels~\cite{chexpert} and modified for higher relevance and better coverage of the possible medical findings. The relevance label is central to \ourmethod{} rather than an auxiliary annotation: it determines whether an incorrect or missing finding constitutes a clinically consequential error. By separating relevance from accuracy, \ourmethod{} distinguishes errors that could affect patient management from minor omissions or deviations that should not render an otherwise useful report inadmissible. This distinction is essential because counting all discrepancies equally would obscure the practical clinical utility of generated reports.

\subsection{Radiologist evaluation} \label{sec:dataset_eval}
Two board-certified radiologists reviewed the DICOM images, the clinical \textsc{Reason}, the original \textsc{Findings}, and the generated reports corresponding to \nlabeled{} studies using the \ourmethod{} rubric. Readers were instructed to evaluate each report against the images; the original \textsc{Findings} was treated as a reference, was evaluated under the same rubric, and was not assumed to be wholly correct. Each report received 24 annotations comprising one relevance and one accuracy label for each category.

\subsection{Automatic metric evaluation}
We evaluated 14 automatic metric variants. These comprised RadCliQ-v0 and RadCliQ-v1, BERTScore, RaTE-Score, semb-score, RadGraph-F1, GREEN, BLEU, CheXbert positive-label F1, four official FORTE components, and a custom pooled FORTE score. CheXbert uncertain labels were treated as not positive, and the 12 report pairs with two empty positive-label sets were assigned an F1 score of one. RaTE-Score was computed using RaTE-NER-DeBERTa, BioLORD-2023-C, and the long affinity matrix. GREEN was computed with the GREEN-RadLlama2-7b checkpoint~\cite{ostmeier2024green} using deterministic decoding. FORTE degree, landmark, feature, and impression F1 scores were computed separately. The custom pooled FORTE score is a category-tagged union of the four components and is not an official FORTE statistic. Component F1 was undefined when no terms in that category were matched; these values were treated as missing rather than assigned a numerical score.

\subsection{Conformal prediction} \label{sec:calibration}
We evaluated Conformal Language Modeling (CLM)~\cite{clm} and the generation stage of Conformal Generative Modeling with Improved Sample Efficiency through Sequential Greedy Filtering (SCOPE-Gen)~\cite{scope-gen}. Both methods use an admissibility function \(\mathcal{A}:\mathcal{Y}\rightarrow\{0,1\}\) that labels a candidate report \(y\) as admissible or inadmissible. The strict CheXbert definition is
\begin{equation} \label{eq:1}
	\mathcal{A}(y) =
	\begin{cases}
		1 & \text{if } \ell_k(y)=\ell_k(y^*)\quad\text{for every }k\in\{1,\ldots,K\},\\
		0 & \text{otherwise,}
	\end{cases}
\end{equation}
where \(y^*\) is the reference report and \(\ell_k\) is the \(k\)-th CheXbert label~\cite{chexbert,bert}. Exact matching produced too few admissible candidates for useful calibration. We therefore used a soft-label definition based on positive-label F1: the binary loss was \(1-\mathrm{F1}\), and the primary loss threshold was 0.6, so reports with \(\mathrm{F1}\geq0.4\) were admissible. The fixed disjoint analysis was repeated at thresholds of 0.5 and 0.6. The alternative \ourmethod{} function labeled a report admissible according to the category-level rule described above.

The labeled random-split analyses used 970 candidates from 97 studies, with 77 studies for calibration and 20 for testing in each 80:20 split. The full-set CheXbert analysis used all 293 studies, with 234 for calibration and 59 for testing. The fixed disjoint analysis calibrated on the 196 studies without radiologist labels and evaluated on all 97 labeled studies; no patient or study appeared in both sets. We evaluated conformal miscoverage levels \(\alpha\in\{0.30,0.45\}\), corresponding to target coverages of 0.70 and 0.55. For SCOPE-Gen, the \(\mathrm{count}\) nonconformity score was prediction-set cardinality, whereas \(\mathrm{max}\) was the maximum included candidate score plus 0.1 for every included report. CLM used both lambda components and a maximum of 20.

Each same-label configuration was evaluated across 100 study-level random calibration/test splits. A run was valid only if it produced a finite prediction set. Coverage and prediction-set size are reported as the mean and standard deviation across valid runs, together with \(n_{\mathrm{valid}}\), the number of valid runs among the 100 attempts. The fixed disjoint analysis used one split and is reported without a variability estimate. Results conditional on a small \(n_{\mathrm{valid}}\) are not estimates over all attempted splits.

\subsection{Statistical analysis}
The study, rather than the individual generated report, was the primary independent sampling unit. The primary metric analysis averaged each automatic metric and the binary \ourmethod{} endpoint over the ten reports within each study and measured their association using Spearman's rank correlation. Spearman correlation was selected because the hypothesis concerned monotonic association and did not require normally distributed metric values. The primary analysis included 97 studies and 970 reports for complete metrics. Metric-specific complete-case sample sizes were 96 studies and 957 reports for FORTE impression F1, 97 and 956 for FORTE feature F1, and 97 and 925 for FORTE landmark F1; all other variants included 97 studies and 970 reports. No values were imputed.

Two-sided \(P\) values were computed for each Spearman correlation, and a family-wise hypothesis-testing alpha of 0.05 was used. Holm's procedure adjusted the 14 primary \(P\) values for multiple comparisons. The 95\% confidence intervals are percentile intervals from 2,000 cluster-bootstrap replicates generated by sampling the 97 study identifiers with replacement and retaining all reports from each sampled study. The report-level CheXbert analysis was exploratory, unadjusted, and did not account for within-study dependence, so the study-level analysis was used for inference. Exact \(P\) values are reported in the text and \cref{tab:metric_corr}. Normality testing was not applicable to the nonparametric rank analysis.

The conformal analyses were descriptive and did not use null-hypothesis tests. Their \(\alpha\) denotes conformal miscoverage and is distinct from the 0.05 hypothesis-testing alpha. Values following \(\pm\) in \cref{tab:results0.3,tab:results0.45} are standard deviations across valid random splits.

\subsection{Ethics}
MIMIC-CXR contains de-identified human clinical data distributed through PhysioNet under credentialed access and a data-use agreement~\cite{mimic-cxr}.

\section{Data availability}
MIMIC-CXR is available through PhysioNet under credentialed access and its data-use agreement~\cite{mimic-cxr}. Derived data, including the generated candidate reports and \ourmethod{} annotations, will be made available after successful publication at PhysioNet, according to the MIMIC-CXR data-use conditions.

\section{Code availability}
The supporing codebase for this study is made publicly available and can be accessed via this link: \href{https://github.com/NikosKont/clarity}{\texttt{github.com/nikoskont/clarity}}.

\section{Author contributions}
N.K. and K.-R.K. conducted the experimental analysis and authored the manuscript. P.K. contributed to the CLARITY method design and labelled reports. A.E. labelled reports. C.B., L.K and M.M. provided feedback on the study design and supported manuscript authorship. S.A. and M.M. provided supervision. All authors reviewed and approved the final manuscript.

\section{Acknowledgements}
This research received no specific grant from any funding agency in the public, commercial, or not-for-profit sectors.

\section{Competing interests}
The authors declare no competing interests.

\newpage
\bibliographystyle{unsrt}
\bibliography{ref}

@article{asgari2025framework,
	title   = {A Framework to Assess Clinical Safety and Hallucination Rates of {LLMs} for Medical Text Summarisation},
	author  = {Asgari, Elham and others},
	journal = {npj Digital Medicine},
	volume  = {8},
	number  = {274},
	year    = {2025},
	pages   = {1-15},
}

@article{bert,
	title   = {{BERT}: {P}re-training of Deep Bidirectional {T}ransformers for Language Understanding},
	author  = {Devlin, Jacob and Chang, Ming-Wei and Lee, Kenton and Toutanova, Kristina},
	journal = {Conference of the North {A}merican Chapter of the Association for Computational Linguistics},
	year    = {2019},
	pages   = {4171-4186},
}

@inproceedings{bertscore,
	title     = {{BERTScore}: Evaluating Text Generation with {BERT}},
	author    = {Zhang, Tianyi and Kishore, Varsha and Wu, Felix and Weinberger, Kilian Q. and Artzi, Yoav},
	booktitle = {International Conference on Learning Representations},
	year      = {2020},
}

@article{chexbert,
	title   = {Combining Automatic Labelers and Expert Annotations for Accurate Radiology Report Labeling using {BERT}},
	author  = {Smit, Akshay and others},
	journal = {Empirical Methods in Natural Language Processing},
	pages   = {1500-1519},
	year    = {2020},
}

@article{chexpert,
	title   = {{CheXpert}: {A} Large Chest Radiograph Dataset with Uncertainty Labels and Expert Comparison},
	journal = {AAAI Conference on Artificial Intelligence},
	author  = {Irvin, Jeremy and others},
	year    = {2019},
	pages   = {590-597},
}

@article{clm,
	title   = {Conformal language modeling},
	author  = {Quach, Victor and others},
	journal = {International Conference on Learning Representations},
	year    = {2024},
	pages   = {1-28},
}

@article{conflvlm,
	title   = {Towards Statistical Factuality Guarantee for Large Vision-Language Models},
	author  = {Li, Zhuohang and others},
	journal = {Conference on Empirical Methods in Natural Language Processing},
	year    = {2025},
	pages   = {11435-11456},
}

@article{jain2021radgraph,
	author  = {Jain, Saahil and others},
	title   = {{RadGraph: Extracting Clinical Entities and Relations from Radiology Reports}},
	journal = {{PhysioNet}},
	year    = {2021},
	note    = {Version 1.0.0},
}

@article{kim2025medical,
	title   = {Medical Hallucination in Foundation Models and Their Impact on Healthcare},
	author  = {Kim, Yubin and others},
	journal = {arXiv:2503.05777},
	year    = {2025},
	pages   = {1-61},
}

@article{laang2023artificial,
	title   = {{Artificial intelligence-supported screen reading versus standard double reading in the Mammography Screening with Artificial Intelligence trial (MASAI): a clinical safety analysis of a randomised, controlled, non-inferiority, single-blinded, screening accuracy study}},
	author  = {L{\aa}ng, Kristina and others},
	journal = {The Lancet Oncology},
	volume  = {24},
	number  = {8},
	pages   = {936-944},
	year    = {2023},
}

@article{li2025automatic,
	title   = {Automatic radiology report generation with deep learning: a comprehensive review of methods and advances},
	volume  = {58},
	number  = {11},
	journal = {Artificial Intelligence Review},
	author  = {Li, Yilin and Kong, Chao and Zhao, Guosheng and Zhao, Zijian},
	year    = {2025},
	pages   = {1-42},
}

@article{li2025forte,
	author  = {Li, Cheng-Yi and others},
	title   = {Towards a Holistic Framework for Multimodal {LLM} in {3D} Brain {CT} Radiology Report Generation},
	journal = {Nature Communications},
	volume  = {16},
	pages   = {2258},
	year    = {2025},
	doi     = {10.1038/s41467-025-57426-0},
}

@article{lin2004rouge,
	title   = {{ROUGE}: {A} package for automatic evaluation of summaries},
	author  = {Lin, Chin-Yew},
	journal = {Text summarization branches out},
	pages   = {74-81},
	year    = {2004},
}

@article{lubeck2025adaptable,
	title   = {Adaptable cardiovascular disease risk prediction from heterogeneous data using large language models},
	author  = {L{\"u}beck, Frederike and others},
	journal = {arXiv:2505.24655},
	pages   = {1-29},
	year    = {2025},
}

@article{maira-2,
	title   = {{MAIRA}-2: Grounded radiology report generation},
	journal = {arXiv:2406.04449},
	pages   = {1-72},
	author  = {Bannur, Shruthi and others},
	year    = {2024},
}

@article{medgemma,
	title   = {{MedGemma} {Technical} {Report}},
	journal = {arXiv:2507.05201},
	author  = {Sellergren, Andrew and others},
	pages   = {1-62},
	year    = {2025},
}

@article{mimic-cxr,
	title   = {{MIMIC}-{CXR}, a de-identified publicly available database of chest radiographs with free-text reports},
	volume  = {6},
	number  = {317},
	journal = {Scientific Data},
	author  = {Johnson, Alistair E. W. and others},
	year    = {2019},
	pages   = {1-8},
}

@inproceedings{ostmeier2024green,
	author    = {Ostmeier, Sophie and others},
	title     = {{GREEN}: Generative Radiology Report Evaluation and Error Notation},
	booktitle = {Findings of the Association for Computational Linguistics: EMNLP 2024},
	pages     = {374--390},
	year      = {2024},
	doi       = {10.18653/v1/2024.findings-emnlp.21},
	url       = {https://aclanthology.org/2024.findings-emnlp.21},
}

@article{papineni2002bleu,
	title   = {{BLEU: A Method for automatic evaluation of machine translation}},
	author  = {Papineni, Kishore and Roukos, Salim and Ward, Todd and Zhu, Wei-Jing},
	journal = {Annual Meeting of the Association for Computational Linguistics},
	pages   = {311-318},
	year    = {2002},
}

@article{scope-gen,
	title   = {Conformal Generative Modeling with Improved Sample Efficiency through Sequential Greedy Filtering},
	author  = {Klaus-Rudolf Kladny and Bernhard Sch{\"o}lkopf and Michael Muehlebach},
	journal = {International Conference on Learning Representations},
	year    = {2025},
	pages   = {1-30},
}

@article{scuq,
	title   = {Semantic Consistency-Based Uncertainty Quantification for Factuality in Radiology Report Generation},
	author  = {Wang, Chenyu and Zhou, Weichao and Ghosh, Shantanu and Batmanghelich, Kayhan and Li, Wenchao},
	journal = {Findings of the Association for Computational Linguistics},
	year    = {2025},
	pages   = {1739-1754},
}

@article{shahrokhi2025conformal,
	title   = {Conformal Prediction Sets for Deep Generative Models via Reduction to Conformal Regression},
	author  = {Shahrokhi, Hooman and Roy, Devjeet Raj and Yan, Yan and Arnaoudova, Venera and Doppa, Janaradhan Rao},
	journal = {Uncertainty in Artificial Intelligence},
	year    = {2025},
	pages   = {1-31},
}

@article{tibshirani2023conformal,
	title   = {{Conformal Prediction}},
	author  = {Tibshirani, Ryan},
	journal = {UC Berkeley, Lecture Notes},
	pages   = {1-15},
	year    = {2023},
}

@article{van2019calibration,
	title   = {{Calibration: the Achilles heel of predictive analytics}},
	author  = {Van Calster, Ben and McLernon, David J. and van Smeden, Maarten and Wynants, Laure and Steyerberg, Ewout W.},
	journal = {BMC Medicine},
	volume  = {17},
	number  = {230},
	pages   = {1-7},
	year    = {2019},
}

@article{yu2023evaluating,
	title    = {Evaluating progress in automatic chest X-ray radiology report generation},
	journal  = {Patterns},
	volume   = {4},
	number   = {9},
	pages    = {100802},
	year     = {2023},
	issn     = {2666-3899},
	author   = {Yu, Feiyang and others},
}

@inproceedings{zhao2024ratescore,
	author    = {Zhao, Weike and others},
	title     = {{RaTEScore}: A Metric for Radiology Report Generation},
	booktitle = {Conference on Empirical Methods in Natural Language Processing},
	pages     = {15004--15019},
	year      = {2024},
	url       = {https://aclanthology.org/2024.emnlp-main.836},
}

\end{document}